\documentclass[sigconf,nonacm]{acmart}
\usepackage{graphicx} 
\usepackage{subcaption}
\usepackage{algorithm}
\usepackage{algorithmicx}
\usepackage{algpseudocode}
\usepackage{hyperref}
\usepackage{tikz}
\usetikzlibrary{arrows,quotes,positioning,shapes,calc, fit, backgrounds}

\usepackage{ulem}
\usepackage{xcolor}

\title{Fun-TSG: A Function-Driven Multivariate Time Series Generator with Variable-Level Anomaly Labeling}
\author{Pierre LOTTE}
\email{pierre.lotte@irit.fr}
\affiliation{%
    \institution{Université de Toulouse, IRIT, CNRS}
    \city{Toulouse}
    \country{France}
}
\author{André PÉNINOU}
\email{andre.peninou@irit.fr}
\affiliation{%
    \institution{Université de Toulouse II - Jean Jaurès, IRIT, CNRS}
    \city{Toulouse}
    \country{France}
}
\author{Olivier TESTE}
\email{olivier.teste@irit.fr}
\affiliation{%
    \institution{Université de Toulouse II - Jean Jaurès, IRIT, CNRS}
    \city{Toulouse}
    \country{France}
}



\begin{document}
\begin{abstract}
Reliable evaluation of anomaly detection methods in multivariate time series remains an open challenge, largely due to the limitations of existing benchmark datasets. Current resources often lack fine-grained anomaly annotations, do not provide explicit inter-variable and temporal dependencies, and offer little insight into the underlying generative mechanisms. These shortcomings hinder the development and rigorous comparison of detection models, especially those targeting interpretable and variable-specific outputs. To address this gap, we introduce Fun-TSG, a fully customizable time series generator designed to support high-quality evaluation of anomaly detection systems. Our tool enables both fully automated generation, based on randomly sampled dependency structures and anomaly types, and manual generation through user-defined equations and anomaly configurations. In both cases, it provides full transparency over the data generation process, including access to ground-truth anomaly labels at the variable and timestamp levels. Fun-TSG supports the creation of diverse, interpretable, and reproducible benchmarking scenarios, enabling fine-grained performance analysis for both classical and modern anomaly detection models.
\end{abstract}

\maketitle

\section{Introduction}

In recent years, the volume of time series data has surged with the widespread deployment of sensors in buildings, vehicles, satellites, medicine, and other smart devices. Leveraging this data can help reduce carbon emissions, improve industrial processes, and anticipate equipment failures. Still, robust and interpretable anomaly detection in multivariate time series remains an open challenge.

Although a growing number of anomaly detection methods have been proposed \cite{dengGraphNeuralNetworkBased2021, limMadSGMMultivariateAnomaly2023, malhotraLSTMbasedEncoderDecoderMultisensor2016, suRobustAnomalyDetection2019}, progress remains hindered by the lack of reliable and informative datasets. Nevertheless, recent studies \cite{liuElephantRoomReliable2025, wuCurrentTimeSeries2021, kimRigorousEvaluationTimeSeries2022} have revealed significant limitations in common datasets and evaluation metrics used for benchmarking, including mislabelled anomalies and unrealistic anomaly patterns. These issues undermine the reliability of performance comparisons across anomaly detection methods.

In response, researchers have developed higher-quality datasets and more informative evaluation metrics. Notably, Liu et al. \cite{liuElephantRoomReliable2025} introduced TSB-AD, a manually curated open-source dataset combining real and synthetic multivariate time series, along with a recommended evaluation metric \cite{paparrizosVolumeSurfaceNew2022}. While this marks progress toward more rigorous benchmarking, key limitations remain: the generative mechanisms are opaque and do not explicitly define inter-variable dependencies, time lags, anomaly types, and contaminated variables.

In addition, recent methods like GDN \cite{dengGraphNeuralNetworkBased2021} aim to localize anomalies in both time and variables, but their accuracy is unverifiable due to the lack of ground truth at the variable level. Thus, while valuable, TSB-AD alone does not suffice for meaningful attribution evaluation. Synthetic generators based on GANs \cite{yoon2019time}, diffusion models \cite{galibFIDEFrequencyInflatedConditional2025}, or transformers \cite{chenSDformerSimilaritydrivenDiscrete2025} offer limited control over temporal and inter-variable dependencies. In contrast, fully configurable tools \cite{wenigTimeEvalBenchmarkingToolkit2022b} offer full control but lack realism and temporal dependencies.

\begin{figure*}[t!]
    \begin{minipage}[c]{0.35\textwidth}
        \centering
        \begin{subfigure}[c]{0.90\textwidth}
            \centering
            \begin{tikzpicture}[
                node distance = 1cm,
                every node/.style={circle, draw, minimum size=0.5cm},
                exo/.style={circle, draw, line width=0.7mm, minimum size=0.5cm},
                background/.style={rectangle, draw=none, fill=blue!05, rounded corners, inner sep=3pt, fit=#1},
                background2/.style={rectangle, draw=none, fill=red!05, rounded corners, inner sep=3pt, fit=#1},
            ]
                \node [exo] (A) {$x_{\bullet, 1}$};
                \node [below=1cm of A] (B) {$x_{\bullet, 0}$};
                \node [exo, right=1.5cm of A] (C) {$x_{\bullet, 4}$};
                \node [below left=.5cm and .5cm of C] (D) {$x_{\bullet, 2}$};
                \node [below right =.8cm and .2cm of C] (E) {$x_{\bullet, 3}$};

                \draw[->] (A) -- (B);
                \draw[->] (C) -- (D);
                \draw[->] (D) to[out=-40, in=-160]  (E);
                \draw[->] (E) to[out=140, in=20] (D);
                \draw[->] (E) edge[loop right] (E);
        
            \end{tikzpicture}
            \caption{}
            \label{fig:graph}
        \end{subfigure}
        \begin{subfigure}[c]{0.90\textwidth}
            \begin{minipage}{0.95\textwidth}
                \begin{align*}
                    x_{i, 0} &= \cos((i-2)x_{i-3, 1} )\\
                    x_{i, 1} &= \frac{\cos(9(i-4))}{\sin(9)}\\
                    x_{i, 2} &= \frac{\cos((i-2) x_{i-2, 4}) + 2x_{i-4, 4} - \frac{x_{i-3, 3}}{4}}{10}\\
                    x_{i, 3} &= \sin(i-3) - \int^{i-1}_{i-3} x_{\bullet, 2} dx_{\bullet, 2} + \frac{x_{i-3, 3}}{2}\\
                    x_{i, 4} &= \sin(6(i-4)) + (3\cos(i-1)-2)^2\\
                \end{align*}
            \end{minipage}
            \caption{}
        \end{subfigure}
    \end{minipage}
    \begin{minipage}[c]{0.60\textwidth}
        \centering
        \begin{subfigure}[c]{0.90\textwidth}
            \begin{minipage}{0.95\textwidth}
                \begin{align*}
                    x_{i, 3} &= \sin(i-3) - \int^{i-1}_{i-3} x_{\bullet, 2} dx_{\bullet, 2} + \frac{x_{i-3, 3}}{\color{red}{5}}
                \end{align*}
            \end{minipage}
            \caption{}
        \end{subfigure}
        \begin{subfigure}[c]{0.99\textwidth}
            \includegraphics[width=\textwidth]{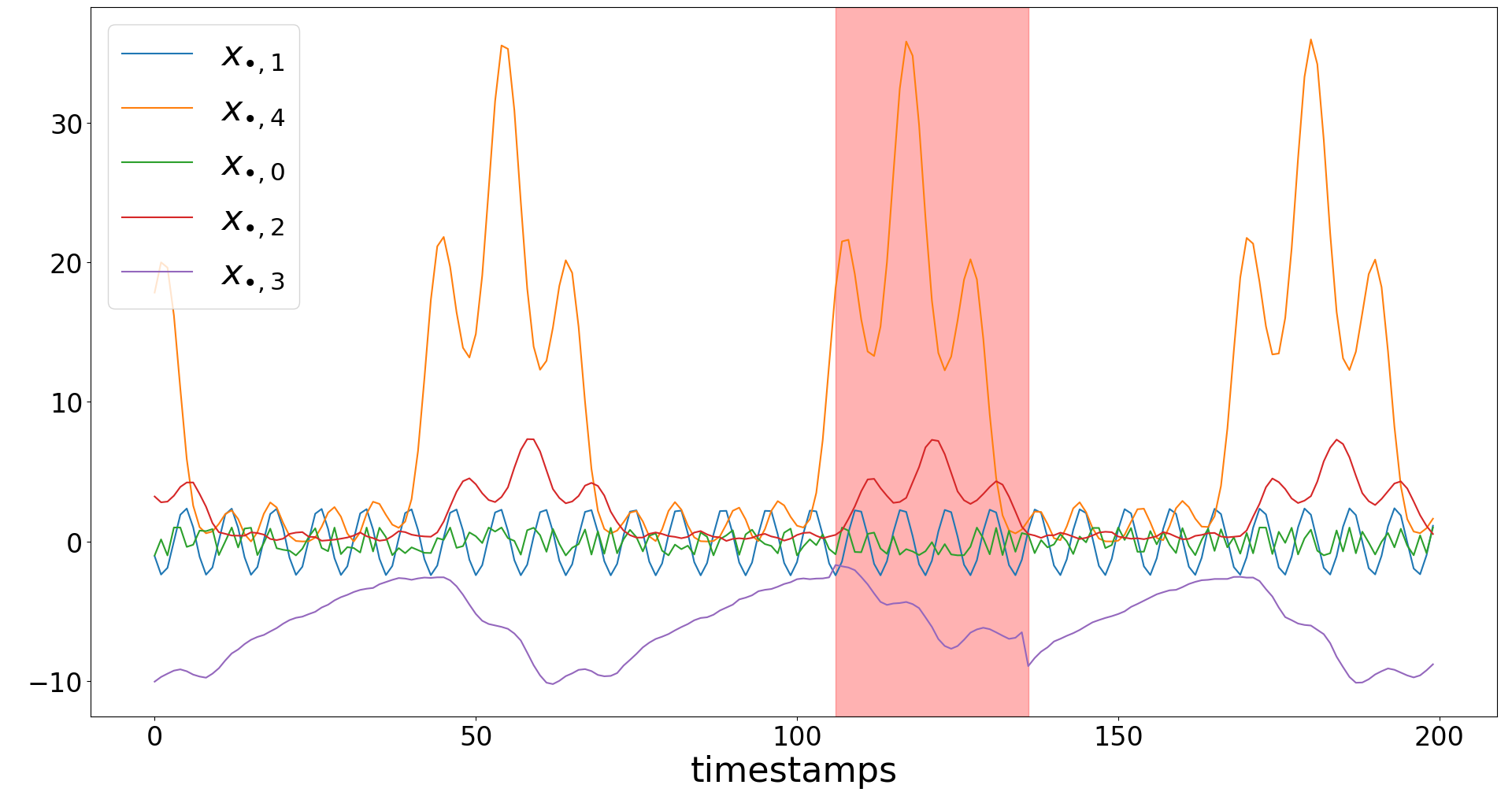}
            \caption{}
        \end{subfigure}
    \end{minipage}
    \caption{
        Illustration of a synthetic multivariate time series generated by \textsc{Fun-TSG}. 
        \textbf{(a)} Dependency graph $\mathcal{G}$ showing directed relationships among five variables. Edges indicate functional dependencies used to generate each variable based on its parents and bold nodes represent exogenous variables. 
        \textbf{(b)} The system of equations we want to use as a generative process.
        \textbf{(c)} The contaminated equation version of the equation of $x_{\bullet, 3}$.
        \textbf{(d)} Corresponding time series plot with five variables. Anomaly injected in $x_{106:136, 3}$ causes deviation (highlighted in red).
    }
    \label{fig:example}
\end{figure*}
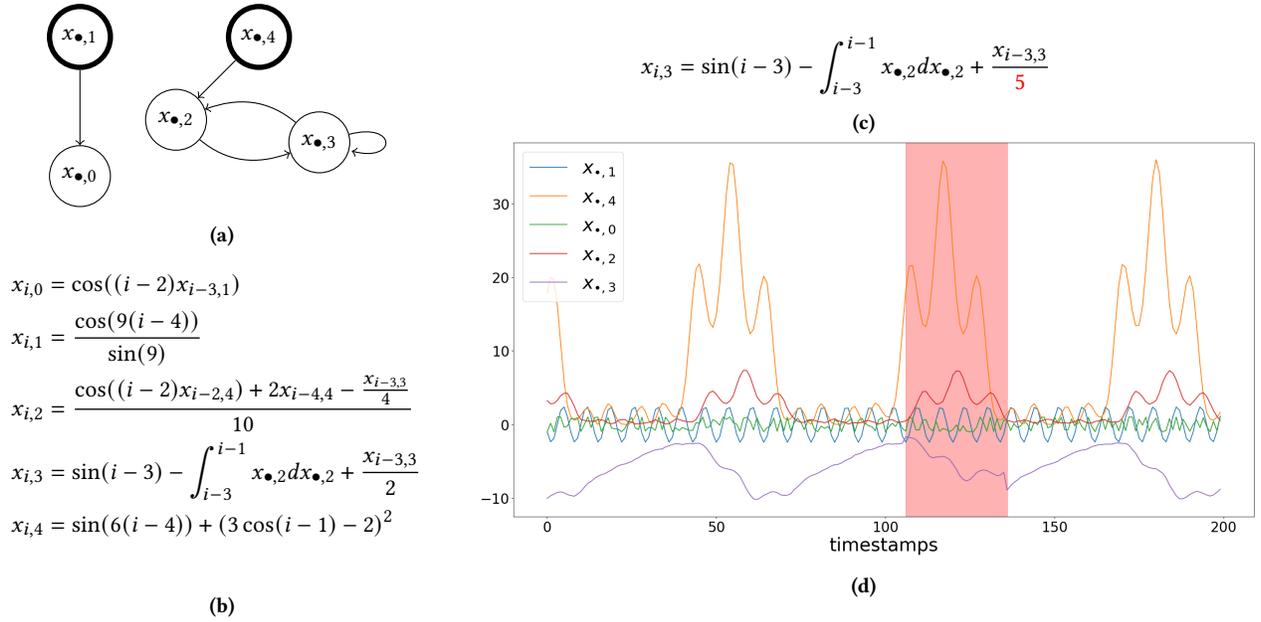

In this context, we introduce Fun-TSG, a new time series generator. It enables customizable generation of multivariate time series with explicit control over temporal and inter-variable dependencies. It also provides full transparency, including ground-truth anomaly annotations at both variable and timestamp levels. This enables precise interpretation and rigorous evaluation of models, including the identification of contaminated variables.

\section{Methodology}

This section presents our data generation methodology, describing its core mechanisms and configurable parameters. Our goal is to demonstrate how the tool can be used to generate multivariate time series with adjustable temporal and inter-variable dependencies, as well as the injection of precisely annotated anomalies of any type.

\subsection{Notations}
\label{sec:notation}


Let $X=(x_{i,j}) \in \mathbb{R}^{T\times d}$ be a multivariate time series of length $T$ with $d$ variables. We denote by $x_{\bullet, j} \in \mathbb{R}^T$ the time series of the $j$-th variable, and by $x_{i, \bullet} \in \mathbb{R}^d$ the $i$-th multivariate observation. The set of all variables is denoted $\mathcal{V} = \{x_{\bullet, j} \mid 0\leq j < d\}$.

A \textit{partial dependency} of variable $x_{\bullet, j}$ on the variable $x_{\bullet, j'}$ occurs when the values of $x_{\bullet, j}$ are at least partially determined by those of $x_{\bullet, j'}$. Let $\Omega_{x_{\bullet, j}}$ denote the set of potentially time-lagged variables upon which $x_{\bullet, j}$ depends. We define $f_{x_{\bullet, j}}$ as the \textit{generative function} that maps $\Omega_{x_{\bullet, j}}$ to $x_{\bullet, j}$, i.e., $x_{\bullet,j} = f_{x_{\bullet, j}}(\Omega_{x_{\bullet, j}})$.

The generative process of multivariate time series is modelled using a directed graph denoted $\mathcal{G} = (\mathcal{V}, \mathcal{E})$, where an edge representing a partial dependency of variable $x_{\bullet, j}$ on $x_{\bullet, j'}$ is denoted $(x_{\bullet, j'}, x_{\bullet, j}) \in \mathcal{E}$, i.e., $x_{\bullet, j'} \in \Omega_{x_{\bullet, j}}$. In $\mathcal{G}$, we call \textit{exogenous variables} elements of $\mathcal{V}$ that satisfy $\Omega_{x_{\bullet, j}} = \emptyset$. If this predicate is not verified for a given element of $\mathcal{V}$, we call it an \textit{endogenous variable}.  

Each value $x_{i, j}$ is assigned a label $y_{i, j} \in \{0,1,2,3\}$, where $y_{i, j} = 1$ indicates an anomaly and $y_{i, j}=0$ indicates normal behavior. The other two values will be explained in Section~\ref{sec:anomaly} for clarity's sake. Following standard definitions \cite{blazquez-garciaReviewOutlierAnomaly2022, chandolaAnomalyDetectionSurvey2009}, we define an observation as abnormal if it deviates from the usual generative behavior of the variable. 

We define a subsequence as a contiguous window in time over a subset of variables. For any $V \subseteq \mathcal{V}$, we denote by $x_{i:i+\ell-1, V}$ a subsequence of length $\ell$ starting at time step $i$ over variables in $V$.

In our setting, a subsequence $x_{i: i+\ell-1, j}$ is considered abnormal if it is produced by a function $g_{x_{i:i+\ell-1, j}}$ that deviates from the standard generative process $f_{x_{\bullet, j}}$ of variable $x_{\bullet, j}$. Consequently, we set $y_{t, j} = 1$ for all $t \in \{i, \dots, i+\ell-1\}$.
An illustrative example of this notation and concepts is shown in Figure~\ref{fig:example}.

Building on this understanding of time series and anomalies, our tool aims to achieve three key objectives. 
\begin{itemize}
\item Model the generative process of the multivariate time series using a directed graph $\mathcal{G}$ that makes inter-variable dependencies explicit and interpretable.
\item Define functions $f_{x_{\bullet, j}}$, either manually or via random generation, that capture the typical variable behavior.
\item Inject anomalies by altering $f_{x_{\bullet, j}}$ to simulate deviations from normal behavior, enabling the modeling of diverse abnormal patterns across arbitrary variable subsequences.
\end{itemize}

\subsection{Automatic generation}

To support broad coverage of behaviors and dependencies, our tool enables fully automated data generation. Users specify high-level parameters ($\mathcal{P}$)--number of variables, communities, maximum indegree in $\mathcal{G}$, contamination ratio, and time series length\footnote{For more information on available parameters and configuration options see: \url{https://gitlab.irit.fr/sig/theses/pierre-lotte/Fun-TSG.git}}. The tool then generates $\mathcal{G}$, assigns functions, and synthesizes the data.

The general algorithm used to generate data is described in algorithm~\ref{alg:automatic}.

\begin{algorithm}
    \caption{Automatic data generation algorithm}
    \label{alg:automatic}
    \begin{algorithmic}
    \State \textbf{Input:} User parameters $\mathcal{P}$
    \end{algorithmic}
    \begin{algorithmic}[1]
    \State $\mathcal{G}\gets$ Generate\_Random\_Graph($\mathcal{P}$) \Comment{See algo.~\ref{alg:graph}}
    \ForAll{$v \in \mathcal{V}$}
        \State $f_{x_{\bullet, v}}\gets$ Generate\_Random\_Function($\mathcal{P}$, $\mathcal{G}$) \Comment{See algo.~\ref{alg:function}}
    \EndFor
    \State Generate\_Random\_Anomalies\_Functions($\mathcal{P}$, $\mathcal{G}$) \Comment{See algo.~\ref{alg:anomalies}}
    \State $x_{0:T_{\text{train}}, \bullet}\gets$ Compute\_Values($\mathcal{G}$)
    \State $x_{T_{\text{train}}:T_{\text{test}}, \bullet}\gets$ Compute\_Anomalies($\mathcal{G}$)

    \State $y\gets$ Compute\_Labels($\mathcal{G}$)
    
    \State \Return{$x_{0:T_{\text{train}}, \bullet}, x_{T_{\text{train}}:T_{\text{test}}, \bullet}, y, \mathcal{G}$}
    \end{algorithmic}
\end{algorithm}

\subsubsection{Graph generation}

We automatically generate a dependency graph to model interactions among the time series variables using Algorithm~\ref{alg:graph}.

Our goal is to closely emulate real-world multivariate time series. In many physical systems, not all variables are causally or statistically dependent. For instance, temperature and energy usage in two distant ground-floor rooms of a smart building are unlikely to exhibit a strong and actionable relationship.

To reflect this structure, we partition variables into distinct communities (Line 2). Some variables may be left unassigned for potential use in inter-community connections.

After initializing the communities, we generate the set of edges $\mathcal{E}$. Each community is partitioned into exogenous variables, which have no incoming edges, and endogenous variables, which depend on others (L.5). For each endogenous variable, we assign a set of parent variables by pseudo-randomly selecting them from the same community, while ensuring that the resulting undirected structure remains intra-community connected (L.8 to 10). 

Finally, if inter-community connections are enabled, we create a small number of links between variables from different communities, either by selecting isolated variables or by randomly sampling across clusters (L.14 to 19). An example of a possible graph generated by our method is shown in Figure~\ref{fig:graph}\footnote{The user parameters used to generate this example are the following: number of communities $= 2$, number of variables $=5$, max indegree $= 4$}.

\begin{algorithm}
    \caption{Graph generation algorithm}
    \label{alg:graph}
    \begin{algorithmic}
    \State \textbf{Input:} User parameters $\mathcal{P}$
    \end{algorithmic}
    \begin{algorithmic}[1]
    \State $\mathcal{V}\gets \{x_{\bullet, j} \mid  {0\le j < \mathcal{P}.d}\}$
    \State $\mathcal{C}\gets$ Assign\_To\_Communities($\mathcal{V}, \mathcal{P}.\text{num\_communities})$
    \State $\mathcal{E}\gets \emptyset$
    
    \Comment{Creates edges within each community}
    \ForAll{$c \in \mathcal{C}$}
        \State $V_{c, \text{exo}}, V_{c, \text{endo}}\gets$ Pick\_Exogenous\_Variables($c$)
        \State Nb\_Edges$\gets$ Random($|c|, |V_{c, \text{endo}}|\times\mathcal{P}.\text{max\_indegree}$)
        \State $\mathcal{E}_c\gets \emptyset$
        \While{$|\mathcal{E}_c| < $ Nb\_Edges}
            \State $\mathcal{E}_c\gets\mathcal{E}_c\cup\{\text{Pick\_Random\_Edge}(c)\}$ 
            \State \Comment{Ensures connectivity of undirected structure}
        \EndWhile
        \State $\mathcal{E}\gets\mathcal{E}\cup\mathcal{E}_c$
    \EndFor

    \Comment{Optional linking between communities}
    \If{$\mathcal{P}.\text{link\_communities}$}
        \For{$\mathcal{P}.\text{nb\_links}$}
            \State $c_1, c_2\gets$Pick\_Distinct\_Communities($\mathcal{C}$)
            \State $\mathcal{E}\gets \mathcal{E}\cup \{\text{Pick\_Random\_Link}(c_1, c_2)\}$
        \EndFor
    \EndIf

    \State \Return{$(\mathcal{V}, \mathcal{E})$}
    \end{algorithmic}
\end{algorithm}

Having constructed the dependency graph, we now generate the functional equations $f_{x_{\bullet, j}}$ for each variable, which determine its values based on its parent set in the graph.

\subsubsection{Function generation}
\label{sec:function}

To populate MTS values, we generate variable-specific functions (see Algorithm~\ref{alg:function}). Our generator builds symbolic expresssions using an extensible set of mathematical operators and functions denoted $\mathcal{F}$ (e.g., logarithm, sine, cosine, exponential, addition, division) to build symbolic expressions.

To define $f_{x_{\bullet, j}}$ we construct a syntax tree whose leaves are constants or elements of $\Omega_{x_{\bullet, j}}$. The tree is built bottom-up by repeatedly merging leaves or subtrees until a single complete expression is obtained.

The merging process proceeds as follows. We initialize a set of operands $O_{x_{\bullet, j}}$ containing elements from $\Omega_{x_{\bullet, j}}$ and random constants (L.1). At each step, we randomly select a function or operator from $\mathcal{F}$ (L.4). If a unary function is selected, we pick one operand from $O_{x_{\bullet, j}}$ (L.3); if a binary operator is selected, we pick two (L.6). The function or operator is applied to the operand(s), and the resulting expression is added back to $O_{x_{\bullet, j}}$ (L.7 or 9). This process is repeated until a single expression remains, representing the complete function $f_{x_{\bullet, j}}$.

However, randomly sampling from $\mathcal{F}$ can produce unbounded or rapidly growing expressions that are not realistic for modeling physical systems. To address this, we regulate the use of functions and operators based on their growth behavior. For example, we reduce the probability of generating expressions composed entirely of exponentials and multiplicative terms.

To regulate the generation of unbounded or rapidly growing expressions, we assign a score to each element of $\mathcal{F}$ that reflects its asymptotic behavior. Positive scores indicate a growth-promoting operations (e.g., exponential), while negative scores indicate attenuating effects (e.g., logarithm). When selecting an operator to apply to an operand from $O_{x_{\bullet, j}}$, we adjust the sampling probabilities based on the cumulative growth score of that operand. For example, if the operand has a high cumulative growth score, we reduce the likelihood of selecting strongly amplifying operators such as exponential or multiplication, and instead favor attenuating operations such as logarithm and division.

\begin{algorithm}
    \caption{Function generation algorithm}
    \label{alg:function}
    \begin{algorithmic}
    \State \textbf{Input:} User parameters $\mathcal{P}$, Set of parent variables $\Omega_{x_{\bullet, j}}$
    \end{algorithmic}
    \begin{algorithmic}[1]
    \State $O_{x_{\bullet, j}}\gets$ Initialize\_Leaves$(\Omega_{x_{\bullet, j}})$ \Comment{Add $\Omega_{x_{\bullet, j}}$ and constants}
    \While{$|O_{x_{\bullet, j}}| > 1$}
        \State Operand$_1$ $\gets $ Pop\_Random$(O_{x_{\bullet, j}})$
        \State Operator $\gets$ Pick\_Random\_Operator$($Operand$_1)$ 
        
        \Comment{Conditioned on cumulative growth score}
        \If{Operator \textbf{is} BINARY}
            \State Operand$_2$ $\gets $ Pop\_Random$(O_{x_{\bullet, j}})$
            \State $O_{x_{\bullet, j}} \gets O_{x_{\bullet, j}} \ \cup$ Merge$($Operand$_1,$ Operator$,$ Operand$_2)$ 
        \Else
            \State $O_{x_{\bullet, j}} \gets O_{x_{\bullet, j}}\ \cup$ Merge$($Operand$_1,$ Operator$)$ 
        \EndIf
    \EndWhile
    \State \Return{$N$} \Comment{Final symbolic expression}
    \end{algorithmic}
\end{algorithm}

By repeating this process for each variable, we generate a complete set of tailored functions that govern normal system dynamics. These will later serve as a baseline for controlled anomaly injection. Optionally, users can add noise to the generated series to simulate measurement imperfections or stochastic effects, enhancing realism.

\subsubsection{Anomaly generation}
\label{sec:anomaly}

As mentioned in Section~\ref{sec:notation}, we define an anomaly as a point or sequence that deviates from the usual behavior. Accordingly, we simulate anomalies by modifying the symbolic expression tree that defines the nominal behavior of each affected variable (see Algo.\ref{alg:anomalies}).

The first step in generating the anomalies is to determine the location of anomalous subsequences. Given a user-specified contamination ratio, the algorithm computes the number of anomalous points $N_{\text{anomalous}}$ to inject (L.1). A random set of subsequence lengths is then sampled such that their total duration equals $N_{\text{anomalous}}$, ensuring the target contamination ratio is met (L.2). By randomly sampling the starting points for each anomaly, the generator creates diverse configurations, including cases where anomalies occur concurrently across different variables.

Before generating anomalous functions, we assign a propagation behavior to each graph edge (L.4). This determines whether, during an anomaly, the child node uses the corrupted or normal values of its parents. This simulates real-world scenarios where systems either propagate errors or remain robust. Propagation behaviors are randomly assigned to encourage diverse interaction dynamics.

Regarding labels, we provide both normal and rich labels. Normal labels have been explained in Section~\ref{sec:notation} and only gives labels $0$ or $1$ in case of normal or abnormal values respectively. In addition to this classical labeling, we provide rich labels flagged with values $2$ and $3$. If $y_{i, j} = 2$ it means that one member of $\Omega_{x_{\bullet, j}}$ is abnormal at timestamp $i$ but its abnormal values are not propagated. However, if $y_{i, j} = 3$, it means that one member of $\Omega_{x_{\bullet, j}}$ is abnormal and the abnormal values are propagated. 

This fine-grained annotation system allow for precise evaluation of methods that provide responsible variables for each anomaly detected.

To create an anomalous function, we randomly apply one of three strategies to modify the symbolic expression tree of a variable. The first strategy inserts a new subtree at a randomly chosen location within the syntax tree of $f_{x_{\bullet, j}}$. The second strategy deletes a subtree and replaces it with a constant to preserve syntactic arity. The third strategy replaces a function or operator node with another of the same arity from $\mathcal{F}$. Each strategy guarantees that the resulting tree is grammatically valid either by inserting constants to preserve arity during deletions or by introducing operators when inserting new subtrees.

The node selection policy varies by transformation type. For deletions, we prioritize nodes near the tree's median depth to avoid injecting anomalies that are either trivially obvious or imperceptibly subtle. By contrast, insertion and replacement strategies select nodes uniformly at random.

\begin{algorithm}
    \caption{Anomalies generation algorithm}
    \label{alg:anomalies}
    \begin{algorithmic}
    \State \textbf{Input:} User parameters $\mathcal{P}$, Dependency graph $\mathcal{G} = (\mathcal{V}, \mathcal{E})$
    \end{algorithmic}
    \begin{algorithmic}[1]
    \State $N_{\text{anomalous}}\gets \mathcal{P}.\text{contamination\_ratio} \times \mathcal{P}.\text{length}$
    \State $T_{\text{anomalies}} \gets $Sort$($Random\_Samples$(0, N_{\text{anomalous}}))$
    \State anomalies $\gets$ empty list
    \State propagation $\gets$ Random\_Propagation$(\mathcal{G})$

    \For{$i \in \{0, \dots, |T_{\text{anomalies}}|-2\}$}
        \State start $\gets$ Random\_Value$(\mathcal{P}.\text{length})$
        \State $t_{\text{start}}, t_{\text{end}} \gets T_{\text{anomalies}}[i]+\text{start}, T_{\text{anomalies}}[i+1] + $ start 
        \State $v \gets$ Random\_Choice$(\mathcal{V})$
        \State $f_{\text{anomalous}} \gets \text{Contaminate}(v.f)$
        \Comment{Returns a modified symbolic tree as described in Section~\ref{sec:anomaly}}
        \State anomalies $\gets \text{Append}(\text{anomalies}, (t_{\text{start}}, t_{\text{end}}, f_{\text{anomalous}}))$
        
    \EndFor

    \State \Return{anomalies, propagation} \Comment{Anomaly list and propagation settings}
    \end{algorithmic}
\end{algorithm}

This completes the automatic generation process, which establishes a synthetic dataset with controlled dependencies, behaviors, and anomalies.

\subsection{Manual generation}

Manual configuration of the data generation process is straightforward. Users specify a generative equation for each variable and, optionally, define alternative functions to simulate anomalies and their temporal positions. The program then generates the time series and returns all associated metadata as in the automatic generation mode.

\section{Availability}

The source code and documentation of Fun-TSG are publicly available at \url{https://gitlab.irit.fr/sig/theses/pierre-lotte/fun-tsg}, example datasets and a default executable file are also available at \url{https://irit.fr/SIG/public/thesis/pierre-lotte/Fun-TSG.zip} to support reproducibility and facilitate adoption by the research community.

\section{Conclusion}

Anomaly detection in multivariate time series is a crucial challenge that remains unsolved. While recent efforts have improved the availability of real-world datasets, existing models still struggle to deliver robust, interpretable, and consistent performance across diverse scenarios. To support progress in this area, we introduced Fun-TSG, a function-driven time series generator that enables both automated and manual generation of synthetic datasets with explicitly defined dependencies and precisely annotated anomalies. By offering full transparency over the generative process, Fun-TSG facilitates rigorous, fine-grained evaluation and interpretation of detection models. We hope that Fun-TSG will serve as a flexible and transparent foundation on which to develop robust and interpretable next-generation anomaly detection systems.

\bibliographystyle{ACM-Reference-Format}
\bibliography{Biblio}

\end{document}